\documentclass[10pt,twocolumn,letterpaper]{article}

\usepackage[pagenumbers]{wacv} 

\usepackage{graphicx}
\usepackage{amsmath}
\usepackage{amssymb}
\usepackage{booktabs}

\usepackage[pagebackref,breaklinks,colorlinks]{hyperref}

\usepackage[capitalize]{cleveref}
\crefname{section}{Sec.}{Secs.}
\Crefname{section}{Section}{Sections}
\Crefname{table}{Table}{Tables}
\crefname{table}{Tab.}{Tabs.}

\def\wacvPaperID{2388} 
\def\confName{WACV}
\def\confYear{2025}

\begin{document}

\title{Direct Coloring for Self-Supervised Enhanced Feature Decoupling}

\author{Salman Mohamadi\\
West Virginia University\\
Morgantown, WV, USA\\
{\tt\small sm0224@mix.wvu.edu}
\and
Gianfranco Doretto\\
West Virginia University\\
Morgantown, WV, USA\\
{\tt\small gianfranco.doretto@mail.wvu.edu}
\and
Donald A. Adjeroh\\
West Virginia University\\
Morgantown, WV, USA\\
{\tt\small donald.adjeroh@mail.wvu.edu}
}
\maketitle

\begin{abstract}
  The success of self-supervised learning (SSL) has been the focus of multiple recent theoretical and empirical  studies, including the role of data augmentation (in feature decoupling) as well as complete and dimensional representation collapse. 
   While complete collapse is well-studied and addressed, dimensional collapse has only gain attention and addressed in recent years mostly using variants of redundancy reduction (aka whitening) techniques. In this paper, we further explore a complementary approach to whitening via feature decoupling for improved representation learning while avoiding representation collapse. 
   In particular, we perform feature decoupling by 
   early promotion of useful features via careful feature coloring.  The coloring technique is developed based on a Bayesian prior of the augmented data, which is inherently encoded for feature decoupling. 
 We show that our proposed framework is complementary to the state-of-the-art techniques, while outperforming both contrastive and recent non-contrastive methods. We also study the different effects of coloring approach to formulate it as a general complementary technique along with other  baselines.
\end{abstract}

\section{Introduction}
\label{sec:intro}

Self-supervised learning (SSL) 
provides state-of-the-art results in unsupervised learning, outperforming deep active learning \cite{mohamadi2022deep} and semi-supervised learning, while rivaling supervised learning under different settings. 
Specifically, the core idea of SSL frameworks is to train a model on properly augmented data \cite{tian2020makes} to accomplish a proxy task (also called pretext task) guided by an appropriate loss function \cite{jing2020self}. Despite the emergence of a variety of techniques, 
a majority of the approaches are based on a first principle \cite{mohamadi2023fussl}, enforcing invariance to the representation of augmented data. Seeing it from the perspective of Information Bottleneck (IB) principle \cite{tishby2000information}, the goal is to learn a representation that is very much informative about the data distribution while un-informative of the augmentation. Recent literature offers a multitude of research on both pretext tasks and loss functions, leading to the emergence and evolution of different sets of frameworks, including contrastive, non-contrastive, clustering-based, and whitening-based approaches \cite{mohamadi2024active}. Though less explored, the augmentation process also has been investigated recently \cite{tian2020contrastive,bai2022directional,wen2021toward,wen2022mechanism}. Recent theoretical investigations along with empirical assessment of the learning process of SSL presented a number of findings regarding the elements behind its tremendous success \cite{mohamadi2023more}. Among them, it is established that augmentation is essential as it helps decouple two sets of features, sparse (useful) features  and dense (less useful) features, leading to learning meaningful representations \cite{wen2021toward,wen2022mechanism} with respect to downstream tasks. The leading argument here is that augmentation decouples these two types of features, as proper augmentation reduces the correlation  between dense features while keeping the correlation between sparse features. 

In essence, proper augmentation encourages learning of useful features (sparse features)  by perturbing mainly the dense features.
In fact, it is mostly feature decoupling that is used to counteract the void created by lack of labels in SSL. Specifically, unlike in supervised learning where the labels guide the learning process toward learning and encoding useful features (as they are shared in samples with the same label), in SSL, learning useful features is due to feature decoupling.
Thus, enhancing feature decoupling can be expected to significantly improve the learning mechanism in SSL. 
However,  we argue that one less noticed downside effect of  the augmentation process seems to be its indirect contribution to representation collapse. Representation collapse is a common phenomena in SSL training process, where essentially the learning process leads to some sort of trivial representation. While complete collapse is the main type  of representation collapse, recently, another type of collapse, namely dimensional collapse, has also been characterized \cite{hua2021feature}. One way to think of the complete collapse is to see it as a special solution to the optimization where the corresponding representation is constant (all features are constant). Dimensional collapse, however, emerges out of highly correlated dimensions in the representation, where dimensions collapse to a single dimension (or potentially much fewer than actual number of dimensions). Complete collapse has been well-addressed by techniques such as careful training protocols \cite{chen2020simple}, asymmetric architectural design and training protocols \cite{grill2020bootstrap,chen2021exploring}.
These in essence inject some variance to avoid having zero variance (complete collapse). In contrast, dimensional collapse is well-addressed in recent work on whitening embedding/ latent space \cite{ermolov2021whitening,zbontar2021barlow} by standardizing some covariance matrix, in order to eliminate high correlation between dimensions of representation.
One potential downside of these set of approaches is that the whitening process could generally limit the capacity of the model \cite{siarohin2018whitening}, especially if used earlier in low level feature learning. In other words, 
if done without a careful attention, whitening 
could hinder the feature decoupling provided  by the augmentation process as it decorrelates the dimensions regardless of its relevance (or otherwise) to the target/desired representation.

In this paper, we present a technique to alleviate this downside of whitening by direct coloring to further enhancing feature decoupling, while promoting a faster learning process. Our proposed approach applies to whitening and non-whitening based approaches. 
We also theoretically and empirically examine how coloring would substantially reduce the chance of complete collapse, the 
primary type of collapse.  
Our key contributions are as follows:
\begin{itemize}
    \item We propose a technique based on coloring transform to further enhance feature decoupling based on augmentation, leading to improved  performance. We also empirically show a faster learning convergence, and discuss the avoidance of complete collapse using constrained optimization.

    \item We develop a direct coloring technique privileged by a Bayesian prior
    that does not require the conventional stage of whitening before coloring, allowing for faster coloring transform on the cross-correlation matrix of some embedding space.

    \item We perform a detailed empirical study, suggesting that while coloring is 
    most effective for whitening-based SSL frameworks, a simple variation (weaker version) also improves some other existing non-whitening baselines.
\end{itemize}
\section{Preliminary and background}
\textbf{{1. SSL}}: 
Enforcing invariance to the representation of augmented views is the driving first principle of 
most existing SSL approaches. 
This core idea has been instantiated via a variety of methods including contrastive approaches \cite{chen2020simple}, non-contrastive approaches \cite{grill2020bootstrap,chen2021exploring}, clustering-based methods \cite{caron2020unsupervised,caron2018deep}, whitening-based techniques \cite{ermolov2021whitening,zbontar2021barlow}, etc. Along with this, there have been parallel efforts 
on improved augmentation protocols \cite{tian2020makes}, sampling strategies \cite{azabou2021mine,wang2020unsupervised,xu2022seed}, and robustness \cite{saha2022backdoor}. 
Studies on representation collapse and theoretical justification of approaches and results have also been considered  \cite{wen2021toward,wen2022mechanism}.  
Augmentation effect is indirectly connected 
with representation collapse. However, {augmentation} is also the main source of feature decoupling \cite{wen2021toward}, orienting the learning toward useful sparse features, similar to the role of labels in supervised learning.
While existing standard augmentation protocols generally aim at useful feature decoupling, not every augmentation protocol leads to the \textit{desired feature decoupling}. That is,
certain augmentations may not necessarily lead to decoupling sparse (useful) and dense (less useful) features \cite{wen2021toward,tian2020makes}. Existing set of augmentation protocols generates views that are easy to associate for human visual perception. Intuitively, underlying useful features are not often perturbed to the point where positive views are visually  dissociated from  each other.

\textbf{Representation collapse:} Let's say for a given sample $x$, the random augmentation function $\tau$ generates two views $x_1$, and $x_2$, and the goal is to train a network $\Phi(.)$ so that the directions of $\Phi(x_1)$ and $\Phi(x_2)$ align \cite{wen2022mechanism}. Technically, we want the optimizer to find a robust  representation for the augmentation effect. However, the optimizer might come up with practically meaningless representations that theoretically fit the optimization objective. Theoretically, one can analyse such cases in terms of variance and covariance of the representation. The most common type of such solutions is when $\Phi(.)$ 
leads to a constant vector, hence the {variance } of the representation is zero \cite{hua2021feature,wen2022mechanism}. 
This is called \textit{complete} or \textit{total collapse}, where the representation collapses to a single dot in the space. 
Another less recognized case is when the coordinates of the representation, $\Phi_i(.)$, are scaled versions of each other, meaning that all of them are aligned. This is the case where the features are highly correlated, and the covariance matrix is far from standard; hence the representation collapses to a single line, also called  \textit{dimensional collapse} \cite{hua2021feature,wen2022mechanism}. The complete collapse 
has been addressed by a variety of techniques that typically add variance to the representation, while dimensional collapse is often  
addressed by standardizing the covariance matrix, practically via a decorrelation process. The latter  involves whitening the latent/embedding space, allowing for decorrelation of the feature dimensions.

\textbf{{2. Whitening and Coloring in SSL and ML}}: 
While whitening and coloring are well-explored concepts in signal processing dating back to decades ago (see \cite{nehorai1982enhancement}), in modern literature of machine learning, they are reinvented with respect to compatibility with the gradient-based learning. From a reductionist perspective, whitening is a technique for decorrelating the covariance matrix, while coloring is to assign/induce desired statistical correlation within the covariance matrix, mainly for Gaussian processes. Specifically, within computer vision problem domains, different techniques for whitening and coloring transform have been used for various purposes, 
for instance, as a generalized form of batch normalization \cite{siarohin2018whitening}, for speeding up the training \cite{siarohin2018whitening}, enhancing  domain generalization and adaptation \cite{roy2019unsupervised,chiu2019understanding}, preserving desired information and statistical characteristics \cite{yoo2019photorealistic}, etc. 
Very recently, whitening has been used for feature decorrelation within SSL frameworks \cite{ermolov2021whitening,zbontar2021barlow,hua2021feature}. The prime idea behind this is that whitening standardizes the covariance matrix, thus resulting in redundancy reduction in the representation, which also helps the network to avoid dimensional collapse. 

Except for some recent methods \cite{ermolov2021whitening,zbontar2021barlow}  which perform whitening on the latent/embedding space, whitening and coloring 
approaches in the literature did not necessarily 
follow this general idea of 
whitening or coloring on the covariance matrix of \textbf{the embedding space}. Rather, they are usually customized for the application at hand. For instance, as a generalization of batch normalization, Siarohin et al \cite{siarohin2018whitening} proposed a whitening and coloring transform performed on the spatial dimensions of a batch of $m$ images, respectively. This  resulted in standardizing the covariance matrix of the given dimension of the batch (whitening), as well as projecting the covariance matrix to that of an arbitrary multivariate Gaussian (coloring). 
However, our coloring technique distinguishes itself from prior techniques in two important ways:
\textbf{ (1)} it performs direct coloring (as discussed later); and \textbf{(2)} the coloring is done in the latent space.

\section{Coloring for enhanced feature decoupling}
\subsection{Description of method}

\begin{figure*}
  \centering
  \includegraphics[scale=.24]{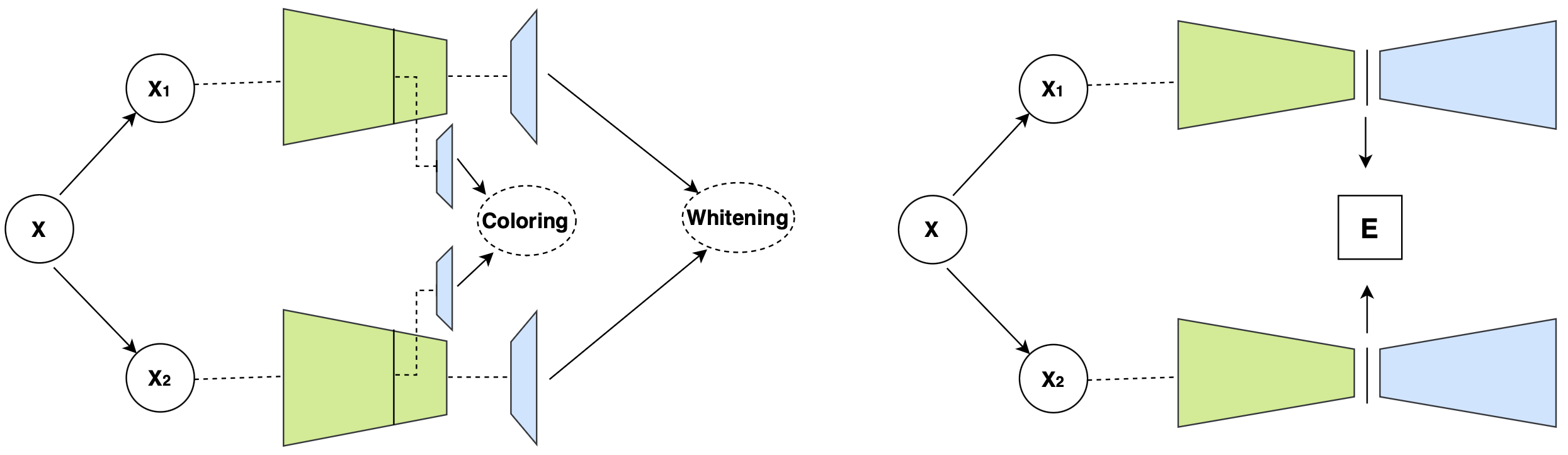}
		\caption{
		Left:Schematic diagram of the proposed framework. For a given sample, two augmented views are generated and fed to the symmetric networks. The two pairs of projectors are used to perform cascade coloring and whitening, respectively. Right: Desired cross-correlation for direct coloring; $E$ is a squared matrix with the same size as the latent space of each of the VAEs.
}
\label{Fig1}
\end{figure*}

Fig. \ref{Fig1} shows a schematic diagram of our proposed framework.  
Our method is conceptually simple and is categorized as a non-contrastive approach, in the sense that it does not require negative views. Similar to some prior work such as \cite{chen2020simple,zbontar2021barlow,ermolov2021whitening}, this framework is also constructed from a pair of symmetric networks, 
where the base design consists of encoders followed by projectors. The basic procedure starts with a random data augmentation function $\tau$ generating two augmented views for each sample image from a batch of samples $X$, namely $X_1$ and $X_2$, to be fed to the pair of symmetric networks. However, our framework distinguishes itself from others by further architectural modification as well as its loss function. The specific core idea is based on enhancing feature decoupling  by inducing meaningful correlation (coloring) between features followed by eliminating  unnecessary redundancy. We will empirically assess how this oriented coloring also diversifies the useful features. We view feature decoupling as the main role of data augmentation, and thus facilitates learning useful sparse features (as opposed to less useful dense features) \cite{wen2021toward,wen2022mechanism}. However, 
we must emphasize that only {proper augmentation} leads to desired feature decoupling, otherwise the learning would not capture plausible general features. Hence, we carefully engineer a coloring transform that allows for effective feature decoupling relying on proper augmentation. This coloring is shortly followed by a decorrelation (aka whitening) process allowing for dimensional collapse avoidance. Assuming the proper coloring (relying on proper data augmentation), the loss function is given as follows:

\begin{equation}
\label{eq1}
\scriptsize
    \mathcal{L}= \mathcal{L}_{W} + \lambda \mathcal{L}_{C}
\end{equation}
where $\mathcal{L_C}$ and $\mathcal{L_W}$ are coloring loss and whitening loss, respectively. Here we have:
\begin{equation}
\label{eq2}
\scriptsize
\begin{split}
     \mathcal{L}_{C} = \sum_i \sum_j(C_{ij}-E_{ij})^2 \:\:\: \text{and } \:\:\:
     \mathcal{L}_{W} = \sum_i(1-W_{ii})^2+\alpha \sum_i \sum_{j\neq i} (W_{ij})^2
     \end{split}
\end{equation}
where, $\lambda$ and $\alpha$ are weighting factors. Also $C_{ij}$ and $W_{ij}$ are elements of cross-correlation matrices computed for the coloring and whitening processes, respectively (see Fig. 1), while $E_{ij}$, is the target cross-correlation matrix used for the desired coloring. In other words, matrix $C$ is the cross-correlation matrix computed between the two output vectors of the coloring projector heads, whereas matrix $W$ is the cross-correlation matrix computed between the outputs of the final projector heads (whitening projector heads). Matrix $E$ is the desired colored cross-correlation matrix, computed as will be explained in the next section. We have:
\begin{equation}
\scriptsize
    C_{ij} \triangleq \frac{\sum_m z_{m,i}^{(1)}z_{m,j}^{(2)}}{\sqrt{\sum_m (z_{m,i}^{(1)})^2} \sqrt{\sum_m (z_{m,j}^{(2)})^2}}
\end{equation}
 where $z^{(1)}$ and $z^{(2)}$ are the normalized outputs of the coloring heads respectively for views $x_1$ and $x_2$, $m$ is 
 the batch size. Similarly, $W{_{ij
 }}$ values are computed from the normalized outputs of the whitening heads. 

A certain layer (close to the output layer) of each network is connected to a projector head intended for coloring the features at this level. The output layer is also connected to a projector head for the decorrelation process, whitening the embedding space.
The coloring transform is performed on the cross-correlation matrix of the outputs of the projectors. The goal is to guide the network to target the desired features,\textbf{ by inducing controlled useful correlation in the cross-correlation matrix}. The redundancy would be reduced in the next stage (the whitening process), by setting the cross-correlation matrix to an identity matrix. In the case of whitening, the diagonal elements are set to 1, encouraging similarity in representation, while the off-diagonal elements are to be close to zero, reducing redundancy in the feature representation \cite{zbontar2021barlow}. 
\subsection{Direct coloring and desired colored cross-correlation}
Consider the standard practice of coloring transform under the Gaussian model for a multivariate Gaussian signal. Here the premise is that the signal (vector of random variables) at hand has a non-identity covariance matrix and the goal is to transform the signal to a new signal with the desired covariance matrix. To this end, first we need to perform the whitening transform, which transforms the multivariate Gaussian signal to a signal with an identity covariance matrix, similar to the ideal white noise. Whitening transform here is essentially a decorrelation by change of basis, and then scaling the principal axes to unit length. After that, the transformed signal will undergo a coloring process in which the data is scaled in desired directions toward desired variances and then rotated. This results in projecting the covariance matrix of the signal to that of an arbitrary multivariate Gaussian signal, hence the covariance matrix of the colored signal will be the desired covariance matrix \cite{hossain2014whitening}. 

However, in SSL literature in general, the purpose of whitening is mainly decorrelation, as was done in 
\cite{ermolov2021whitening,zbontar2021barlow}. We also follow this idea, aiming at instantiating the colored signal as the one with the desired cross-correlation. To this end, we deviate from the general procedure of coloring in which coloring is preceded by whitening, and introduce direct coloring relying on some Bayesian prior (as discussed in the next section) which is {compatible with gradient based learning}, while conceptually easier to perform. 
We perform direct coloring by  projecting the (estimated) cross-correlation matrix 
to the desired cross-correlation via a gradient-based learning process minimizing the following functional:
\begin{equation}
\label{eq3}
\begin{split}
\scriptsize
     &\mathcal{L}_{C} = \sum_i \sum_j(C_{ij}-E_{ij})^2\\
     \end{split}
\end{equation}
where $E_{ij}$ are the elements of the desired cross-correlation matrix, whereas $C_{ij}$ are elements of the cross-correlation matrix computed from two embedding vectors. 

We can identify two advantages of direct coloring:
\begin{enumerate}
    \item The process is  
    computationally faster as it does not require us to initially project the elements of $C$ to the identity matrix before setting them to desired values.
    \item It does not assume a multivariate Gaussian, and in fact, the nature of the desired cross-correlation depends heavily on the effect of the argumentation.
\end{enumerate}

Proper or standard augmentation is an integral part of any SSL framework as it decouples features properly, allowing for learing useful features.
We build upon this to further enhance the feature decoupling. To this end, the desired colored cross-correlation matrix is computed from augmented data by further decorrelating the dense features while keeping the correlation between sparse features. Specifically, a pair of variational autoencoders (VAE) are trained separately on pairs of augmented views $x_1$ and $x_2$ for sample image $x$ under standard augmentation protocols. Then $E$, the cross-correlation between the normalized latent vectors 
from the VAEs are computed as the desired cross-correlation. See Fig. \ref{Fig1} (Right). The specific settings and scenarios are further discussed under experiments and ablation study. 
Elements of $E$ are computed along the whole dataset samples ($n$) as follows:
\begin{equation}
\scriptsize
    E_{ij} \triangleq \frac{\sum_n z_{n,i}^{(1)}z_{n,j}^{(2)}}{\sqrt{\sum_n (z_{n,i}^{(1)})^2} \sqrt{\sum_n (z_{n,j}^{(2)})^2}}
\end{equation}
 where $z^{(1)}$ and $z^{(2)}$ are the latent vectors of the top and bottom VAEs in Fig. \ref{Fig1} (Right).

\subsection{Maximum A Posteriori (MAP) analysis}
Here we want to demonstrate that the proposed loss function based on a Bayesian prior (desired target cross-correlation) is a solution to the Maximum A Posteriori (MAP) estimation, specifically considering the prior and likelihood components in this work. The MAP estimation aims to find the most probable model parameters given both the data and some prior knowledge i.e., it combines a prior probability ${p(\Theta)}$ and a likelihood ${p(X|\Theta)}$ to find the best model parameters ${\Theta}$:
\begin{equation}
\label{eq11}
\scriptsize
\Theta_{MAP} = argmax_{\Theta}[p(\Theta)p(X|\Theta)]
\end{equation}
here we have a prior in the form of the target colored cross-correlation matrix $\mathbf{E}$ from VAEs, which specifies the desired correlation structure between the augmented views. Hence the prior is Gaussian distribution with mean ${E}$ and a certain variance $\sigma^2$ as follows:
\begin{equation}
\label{eq12}
\scriptsize
p(\Theta) = \mathcal{N}(\Theta|E,\sigma^2)
\end{equation}

Moreover, the likelihood is composed of two terms: the coloring likelihood $p_{color}(X | \Theta)$ and the whitening likelihood $p_{whiten}(X | \Theta)$, which capture the respective effects of the coloring and whitening processes on the learned representations. These likelihoods are defined as follows: 
\begin{equation}
\label{eq13}
\scriptsize
p_{color}(X | \Theta) = \prod_i \prod_j \mathcal{N}(C_{ij}|E_{ij},\sigma^2)
\end{equation}

\begin{equation}
\label{eq14}
\scriptsize
    p_{whiten}(X | \Theta) = \prod_i \mathcal{N}(W_{ii}|1,\sigma^2). \prod_{i\neq j} \mathcal{N}(W_{ij}|0,\sigma^2).
\end{equation}
Thus,  we have:
\begin{equation}
\label{eq15}
\scriptsize
\begin{split}
    & \Theta_{MAP} = argmax_\Theta[ \mathcal{N}(\Theta|E,\sigma^2).\prod_i \prod_j \mathcal{N}(C_{ij}|E_{ij},\sigma^2).\\
    & \prod_i \mathcal{N}(W_{ii}|1,\sigma^2). \prod_{i!=j} \mathcal{N}(W_{ij}|0,\sigma^2)]\\
    \end{split}
\end{equation}
Since maximizing the product of probabilities is equivalent to minimizing the negative logarithm of the product, the MAP objective can be expressed in terms of our loss function:
\begin{equation}
\label{eq16}
\scriptsize
\begin{split}
    & \Theta_{MAP} = argmin_\Theta[ -\log \mathcal{N}(\Theta|E,\sigma^2) - \log \prod_i \prod_j \mathcal{N}(C_{ij}|E_{ij},\sigma^2)\\
    & - \log \prod_i \mathcal{N}(W_{ii}|1,\sigma^2). \prod_{i\neq j} \mathcal{N}(W_{ij}|0,\sigma^2)]\\
    \end{split}
\end{equation}
Comparing this expression to our loss function $\mathcal{L}$, we can see that it aligns with the terms in Equation \ref{eq1} and \ref{eq2},i.e., joint optimization of whitening and Bayesian coloring  terms in our loss function reflects the MAP estimation with a Gaussian prior (see supplementary for more analysis).

\section{Experiments and results}
In this section we present the experimental settings as well as empirical results in order to  assess the effectiveness and generality of our proposed  approach. We use  datasets at different scales, and show results for applying our approach on different downstream tasks.

\subsection{Datasets and Baselines}
\label{sec:datasets_and_baselines}
The main part of the experiments is performed on ImageNet dataset \cite{deng2009imagenet}, under linear evaluation on ImageNet as well as transfer learning on smaller datasets for classification task. However, we also assess the approach on detection and segmentation tasks with different datasets. The experimental results are mainly obtained by building on Solo-Learn \cite{da2022solo}, a recently developed open access library of visual SSL approaches. Solo-Learn \cite{da2022solo} provides an implementation of the existing baselines that we compared against in this section.

\textbf{Datasets:} In this study we make use of ImageNet \cite{deng2009imagenet}, CIFAR10/100 \cite{krizhevsky2009learning}, Tiny ImageNet \cite{le2015tiny}, as well as VOC0712 \cite{everingham2009pascal} and COCO \cite{lin2014microsoft}. In our ablation study we use a separate dataset, ImageNet-100.


\textbf{Baselines:} For comparison, we contrast our framework to different classes of the recent baselines, including contrastive, non-contrastive, clustering-based, and whitening (aka redundancy reduction) baselines, as well as baselines primarily based on vision transformers. These baselines include SimCLR \cite{chen2020simple}, BYOL \cite{grill2020bootstrap}, SimSiam \cite{chen2021exploring}, SwAV \cite{caron2020unsupervised}, Barlow-Twins (BT) \cite{zbontar2021barlow},  Whitening-MSE  (W-MSE) with $d=4$ \cite{ermolov2021whitening}, and DINO \cite{caron2021emerging}.

\subsection{Experimental setting}
\subsubsection{Architecture:} 
ResNet18 and ResNet50 \cite{he2016deep} are used for encoder architecture, except in both cases the last layer is replaced with a three layer projector (we call it whitening projector) as described in \cite{zbontar2021barlow}. The last layer of the projector is the output with size 2048. A set of identical projector heads, architecturally similar to the whitening projector are used for the direct coloring. Specifically the layer 16 and 46 of ResNet18 and ResNet50 pass through average pooling layer and then fed to the coloring heads (as shown in Fig. 1), encouraging the feature decoupling in training process for the former layers. The architecture of the VAE 
is based on ResNet18 or ResNet50, and that determines 
the number of the layer that is connected to the coloring head. For instance, in the case of ResNet18, the corresponding VAE to generate desired colored matrix, is made of an encoder consisting of the first 16 layers of ResNet18, a latent space of the same size as the coloring  head output, and a decoder with the same size as the encoder. More detail is available in the supplementary.

\subsubsection{Augmentation protocol:} For a given sample $x$, two augmented views $x_1$ and $x_2$ are generated using augmentation protocols. Regarding the very recent literature, there are two sets of augmentation protocols, a set of standard augmentation protocols \cite{chen2020simple,ermolov2021whitening,zbontar2021barlow}, as well as a set of heavy augmentation protocols \cite{bai2022directional}. Most of the baselines use standard augmentation protocols. We also use standard protocols for the main experiments. In both cases, a set of augmentation techniques are performed via a random process $\tau$. Specifically, for standard augmentation protocols on all datasets we follow the specification 
in \cite{chen2020simple} which include random mirroring, random crop, random color jittering and gray-scaling, and random aspect ratio re-arrangement.

\subsubsection{Implementation details:} 
\label{sec413}
Optimization of all experiments including pre-training and evaluation under linear setting and transfer learning has been performed using Adam optimizer \cite{kingma2014adam}. Transfer learning using ImageNet pre-training of ResNet50 on CIFAR10/100 is based on standard settings available in \cite{chen2020simple}. The size of output of projection heads (both whitening and coloring heads) for ImageNet dataset is 2048, the same as the size of the latent space of VAEs, whereas for CIFAR10/100 and Tiny ImageNet we follow the details in \cite{ermolov2021whitening}.  Augmentation protocols is standard unless otherwise specified. The pre-training is performed for 1000 epochs consistently along all experiments. The value of $\lambda$ in the loss function (\ref{eq1}) is static, and set to $0.05$, however, in the supplementary material, there is a range of experiments with dynamic values for $\lambda$, e.g., decreasing over the number of epochs. Learning rate and other hyperparameters for ImageNet dataset are same as in \cite{zbontar2021barlow}. Learning rate for CIFAR10 and CIFAR100 is set to $3 \times 10^{-3}$; whereas for Tiny ImageNet and VOC0712, the learning rate is set to $2 \times 10^{-3}$. The weight decay is set to $5 \times 10^{-6}$. All other baselines we follow the latest settings presented by \cite{da2022solo}.

\textbf{Direct coloring:} 
Direct coloring is performed using a desired pre-calculated cross-correlation matrix as the target of coloring the embedding of the coloring heads. The coloring heads are made of three linear layers, first two of which are each followed by batch normalization and ReLU, whereas the output layer is fully connected layer of size 2048. The coloring loss measures the difference between the cross-correlation computed from the outputs of the coloring heads, and the target cross-correlation. In case of ResNet18 as the base encoder, layer number 16 is connected to the coloring head, whereas in case of ResNet50 layer number 46 is connected to its corresponding coloring head. Note that a range of experiments regarding the optimum layer are presented later under ablation study. The target cross-correlation for coloring is computed between the latent spaces of two VAEs with the encoder/decoder same size 
the layer connected to the coloring head, i.e., Layer 16 and 46 for ResNet18 and ResNet50, respectively. The target cross-correlation for desired coloring is investigated further in ablation study. 

\textbf{Whitening:} Whitening is performed on the output of the whitening heads, as a standard practice of decorrelation of highly correlated features. The whitening head is architecturally similar to the coloring head, whereas it replaces the last layer of the ResNet architecture. The hyperparameter $\alpha$ is set to $0.01$, trading off between the diagonal and off-diagonal terms in the whitening loss function. 
The elements of $W$, similar to $C$, fall between -1 and 1, representing a spectrum of correlation (positive), no-correlation (zero) and anti-correlation (negative).

\subsection{Evaluation settings}
Standard evaluation is performed under linear and transfer learning settings following the details of former baselines \cite{ermolov2021whitening,zbontar2021barlow,grill2020bootstrap}. For linear evaluation, the common procedure is to remove the projector heads and train a linear classifier placed on top of one of the fixed encoders under the supervised setting over the source data (used for pre-training). 
For transfer learning purposes (classification, detection, and segmentation), the same standard procedure is performed except that the supervised training and evaluation is performed on the target data. Following 
\cite{zbontar2021barlow}, we also perform linear and transfer learning evaluation, except here we have four heads to remove, both whitening and coloring projector heads. In linear evaluation a whitening head is replaced with the linear classifier (a fully connected layer followed by a softmax) for the evaluation process. The learning rate for linear and transfer learning evaluation consistently starts with $10^{-3}$ and exponentially decays to $10^{-6}$ for some 500 epochs. 

\textbf{Linear evaluation:}
Linear evaluation on classification task is performed on ImageNet, Tiny ImageNet, CIFAR10, and CIFAR100. The evaluation consists of 500 epochs of supervised training (on labeled data) and then 
testing. Note that the encoder is fixed and only linear classifier undergoes training.

\textbf{Transfer learning:} Transfer learning consisted of pre-training on ImageNet dataset and evaluation on other datasets, as presented in the results. Accordingly, on classification task, transfer learning was performed with CIFAR10 and CIFAR100, whereas on detection and segmentation task it is performed on VOC0712 and COCO respectively. 
Similar to the case of linear evaluation, the supervised training is performed for 500 epochs before testing the performance. 
\subsection{Results and comparison} 
In this section the results for different datasets, tasks, and learning paradigms are presented. Following is respectively the classification  results with ImageNet, CIFAR1/100, and Tiny ImageNet, as well as object detection results with VOC0712 and segmentation results with COCO.  

\subsubsection{Linear evaluation with ImageNet}
The evaluation on classification task, has been performed on ImageNet. The results are presented in Table 1, evaluating multiple  baselines under 100, 400, and 1000 epochs of pre-training before supervised linear evaluation. The results show that coloring speeds up the convergence of training. 
At 1000 epochs, our approach offers $0.8\%$ improvement over the former best result. We note that, given the difficulty of this challenge, this magnitude of improvement has been difficult to achieve on this problem. This is evident from the relative difference in performance for the prior baselines, as shown in the table.

\begin{table*}
  \centering
  \scriptsize
  \begin{tabular}{p{1.8cm} p{1.5cm}|p{1.5cm} p{1.1cm}|p{1.5cm} p{1.5cm}p{1.5cm} p{2cm}}
    \toprule
    Framework   & \multicolumn{3}{c}{ImageNet} & CIFAR10   & CIFAR100 & Tiny-ImgNet    \\
    \cmidrule(r){2-7}
        & 100   & 400 & 1000 & & & \\
    \cmidrule(r){1-7}
     SimCLR  &66.6 & 70  & 71.06 &91.03 & 66.48  &  49.11  \\
      BYOL  &   68.4 & \textbf{73.1 }&   {74.6 } &  92.39  & \textbf{70.81} &  \textbf{51.05 } \\
       SwAV  &  66.5  & 70.8 & 72.6 &  90.06  & 65.09 & 48.89\\
       SimSiam  & 67.9  & 70.8 & 71.6  &  90.81 &  66.19&  49.85 \\
    W-MSE4  & \textbf{69.5} & 72.6  & 73.7 & 89.27  & 62.20  & 49.51 \\
   {B-Twins}  &  67.4  &  71.4  & 73.6  &  \textbf{92.45}  & 70.51   & 50.11\\
   DINO (RN50)    &  68.1    &   72.4   & \textbf{75.3}  &  {90.46}  & 67.01   & 48.54\\
   \hline
   Ours   & \textbf{69.6}   &  \textbf{73.2}   & \textbf{76.1} &  \textbf{93.47}   & \textbf{72.52}    &  50.27 \\
   \bottomrule
  \end{tabular}
  \caption{  \scriptsize
  Top-1 linear classification accuracy for ImageNet using ResNet50 pre-trained on ImageNet under 100, 400, and 1000 epochs. SwAV reproduction is without multi-crop technique. 
  Our method converges faster in terms of number of epochs, while also providing higher accuracy.  Top-1 linear classification accuracy for CIFAR10, CIFAR100, and Tiny ImageNet, all using ResNet18.}
     \label{table1}
\end{table*}


\begin{table}
  \centering
  \scriptsize
  \begin{tabular}{p{1.4cm} p{2.5cm}|p{2.5cm}}
    \toprule
    Framework   & \multicolumn{2}{c}{Transfer Learning (ImageNet pre-training)}   \\
    \cmidrule(r){2-3}
        & CIFAR10   & CIFAR100 \\
    \cmidrule(r){1-3}
     SimCLR  & 97.52 & 85.41      \\
      BYOL  & \textbf{97.91}   & {86.21}      \\
       SwAV  &  97.53  & 84.19 \\
       SimSiam  &  97.11 &  85.27   \\
    W-MSE4  &  97.02  &  \textbf{86.31 } \\
   {B-Twins}  &  97.29  &  85.03   \\
   \hline
   Ours   &   \textbf{98.19}  &  \textbf{86.80}     \\
    \bottomrule
  \end{tabular}
  \caption{ \scriptsize
  Top-1 transfer learning classification accuracy, pre-trained on ImageNet, and fine-tuned on CIFAR10, and CIAFAR100.}
    \label{table3}
\end{table}

  
\subsubsection{CIFAR10 and CIFAR100, and Tiny ImageNet}
Linear and transfer learning (pre-trained on ImageNet) evaluation with three datasets, CIFAR10, CIFAR100 and Tiny ImageNet on classification task are presented in Table 2 and Table 3 respectively. In case of linear evaluation, our approach slightly outperformed  state-of-the-art on CIFAR10 and CIFAR100, offering respectively $1.02\%$ and $1.71\%$ improvements over the former best result, whereas it remains competitive (second best) with the state-of-the-art on Tiny ImageNet. 

Transfer learning has also been performed with CIFAR10 and CIFAR100, as presented in Table 3. In both cases, the pre-training was performed on ImageNet while fine-tuning with the CIFAR10/100. Our approach slightly outperformed the state-of-the-art, offering $0.28\%$ and $0.49\%$ improvement, respectively, on CIFAR10 and CIFAR100.

\subsubsection{Transfer learning with VOC0712 (Detection Task) and COCO (Segmentation Task)}
Transfer learning with VOC0712 is performed on object detection. We follow setting in MoCo \cite{he2020momentum}, finetuning the encoder. The results in Table 4 show that on AP$_{50}$ setting our method slightly outperformed former baselines, whereas on AP$_{75}$ and AP$_{100}$ is either on par or very competitive with the former baselines. Transfer learning results with COCO on segmentation task, in Table 4, indicate the competitiveness of our direct coloring technique.
\begin{table}
  \centering
  \scriptsize
  \begin{tabular}{p{1.1cm} p{1.cm}|p{1.cm}| p{1.5cm}|p{1cm}|p{1cm}}
    \toprule
    Framework   & \multicolumn{3}{c}{VOC0712} &  \multicolumn{2}{c}{COCO} \\
    \cmidrule(r){2-6}
        & AP$_{100}$   & AP$_{75}$ & AP$_{50}$ & AP$_{100}$ & AP$_{50}$\\
    \cmidrule(r){1-6}
       SwAV  &  56.1  & 62.7 & 82.6 & 33.8 & 55.2 \\
       SimSiam  & \textbf{ 57} & \textbf{63.7} &  82.4 &  34.4 &  56.0 \\
   {B-Twins}  & 56.8  &  63.4  & 82.6 & 34.3 &   56.0\\
   \hline
  Ours   &   56.6  &  \textbf{63.7 }  & \textbf{82.9} & 34.4 & \textbf{56.9} \\
    \bottomrule
  \end{tabular}
  \caption{  \scriptsize
  Transfer learning on object detection task with VOC0712 (using Faster R-CNN), and segmentation task with COCO (using Mask R-CNN). Results for other baselines are taken from \cite{zbontar2021barlow}. Direct coloring performs either on par or better than the state-of-the-art in both tasks. }
  \label{table4}
\end{table}


 \label{sec:computeCost}
\section{Ablation Study}
We performed ablation study to assess contribution offered by different aspects of this framework. Multiple scenarios including scenarios on location of coloring heads, alternative approaches to computing desired coloring matrix, and alternative/simpler architectures were considered. Other scenarios including direct coloring for other methods, and range of $\lambda$ are presented in supplementary materials. The experiments are performed with a separate dataset, ImageNet-100 on classification task, using ResNet18 evaluated after 500 epochs of pre-training. Note that regarding other baselines with direct coloring technique, empirical evidence is presented in supplementary.

\subsection{Presence and location of coloring head }
We assess the effect of the presence, and location of the coloring heads.
Given the standard setting presented for the framework, the coloring heads are connected to layer 16 of ResNet18, with hidden layer dimension of 2048. Under this setting, the baseline performance on classification task is a top-1 accuracy of $80.93\%$. Removing the coloring heads and corresponding optimization terms from the loss function, the framework is reduced to BT, and the accuracy descends to $79.69\%$. However, with the current output size of 2048, there is a $1.24\%$ improvement with the coloring heads.
Next we assess the performance with coloring head connected to different layers.

\textbf{Coloring at layer 10:}
When the coloring heads are connected to layer number 10, the top-1 accuracy descends to $79.97\%$ (from $80.93\%$). We suspect that this 
drop is mainly because layers 11-16 are no longer under coloring process. Here, the desired coloring cross-correlation is computed from a VAE with both encoder and decoder of size 10, same as the location of coloring heads.

\textbf{Coloring at layer 17:} When the coloring heads are connected to the same layer as the whitening heads, layer number 17, the performance drops to $78.81\%$. Roughly speaking, coloring could be seen as the opposite of whitening. Accordingly we hypothesize that what actually happens in case of parallel whitening and coloring heads, is that the result follows the superposition principle, meaning that coloring is cancelled out as the weight factor $\lambda$ in Equation 1, gives less weight to coloring. 

\subsection{Projector Dimension}
The experiments with a range of different projector dimensionality indicate the sensitivity of the method to this factor, as the top-1 accuracy with output size 512, 1024, 2048, and 4096 are respectively $74.31\%$, $77.15\%$, $80.93\%$, and $81.66\%$. This is a similar behavior to that of Barlow-Twins.
\subsection{Desired coloring matrix}
We briefly assess the case in which coloring matrix is computed from a pair of autoencoder, instead of VAEs. The performance degrades to $80.21\%$. The slightly better performance of VAEs might be due to the fact that the latent space of VAE is regularized, which allows for more robust cross-correlation in terms of variance of the elements.  

\subsection{Less computation with auto-correlation}
We consider an architecturally simpler instantiation of the direct coloring in which, the cross-correlation is replaced with auto-correlation, as shown in Fig. 3. The idea turned out to be effective. As shown in Fig. 3, the the framework consists of only one network, in which the encoder undergoes the coloring before passing its output to whitening heads. Specifically, layer 16 of the encoder is connected to a coloring head where the auto-correlation of its output is becoming colored. Similarly in case of whitening, the auto-correlation of the output is becoming whitened. The desired coloring matrix is also an auto-correlation matrix computed from the latent space of one VAE. Amazingly, while the computational complexity reduces by half, the performance only slightly drops to $80.64\%$. Corresponding loss function of coloring and whitening process as well as the total loss is presented in the supplementary material.

\begin{figure}
\label{Fig3}
  \centering
  \includegraphics[scale=.6]{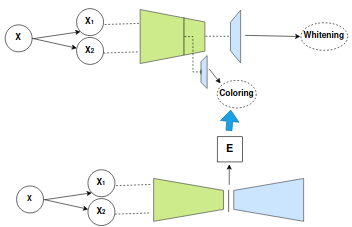}
		\caption{
		Simpler architecture with auto-correlation instead of cross-correlation.
}
\end{figure}

\subsection{On the avoidance of complete collapse}
Let's consider direct coloring a rather general technique for any SSL framework. Here, in terms of loss function, we have two terms, first term being the original loss of the framework to be ungraded denoted as $f$, plus a second term corresponding to the direct coloring, hence we have $\mathcal{L}=f+\lambda g$. In essence, direct coloring in this setting would be seen as a constrained optimization problem in terms of Lagrange multipliers. The first term alone, $f$, would be solved by a trivial solution, complete collapse. However, this is subjected to the second term, $g$, as the coloring constraint. We 
observe that  direct coloring would substantially reduce the chance of complete collapse, because to find the solution to this constrained optimization problem, the optimizer looks for points where the gradient vector of $f$ and $g$ are parallel to each other. Since complete collapse is only a solution to $f$ alone and certainly not a solution to $g$, if one chooses a proper $\lambda$, one can substantially avoid the complete collapse.
Further theoretical and empirical evidence on this as well as empirical evidence on other baselines with direct coloring technique is presented in supplementary .


\vspace{-.25cm}
\section{Conclusion and Future Direction}
In this paper, we analyzed the general setting of SSL, the role of the augmentation process, and trivial solutions to the SSL problem. We extended the study to revisit the core principle underlying state-of-the-art approaches. Building upon this, we presented a new framework for SSL, based on direct coloring, which improved the performance, sped-up learning convergence, and reduced the chance of complete collapse. The key foundation 
is the idea of direct coloring, however, we also considered direct coloring as a general technique that can be used with existing baselines. 
Empirical assessment on multiple datasets and 
three downstream tasks show the effectiveness of the proposed framework. We leave the generalization of direct coloring technique in terms of multi-stage coloring as a future direction.

{\small
\bibliographystyle{ieee_fullname}
\bibliography{egbib}
}

\clearpage
\appendix

\section{ $\lambda$}

\subsection{Sensitivity to $\lambda$}
A range of experiments on ImageNet-100 with different values for $\lambda$ is presented here. Note that the value for $\alpha$ is set to $10^{-2}$ for all experiments and the pre-training was performed only for 250 epochs. The top-1 accuracy for different values of $\lambda$ is depicted in Fig. 1. As it is presented, for values less than $0.03$ down to $0.005$, the accuracy drops somewhat steadily. However, for values greater than $0.05$ the accuracy degradation is sharper. In fact with large values of $\lambda$, dimensional collapse is probable.

\begin{figure}
\label{Fig3}
  \centering
  \includegraphics[scale=.12]{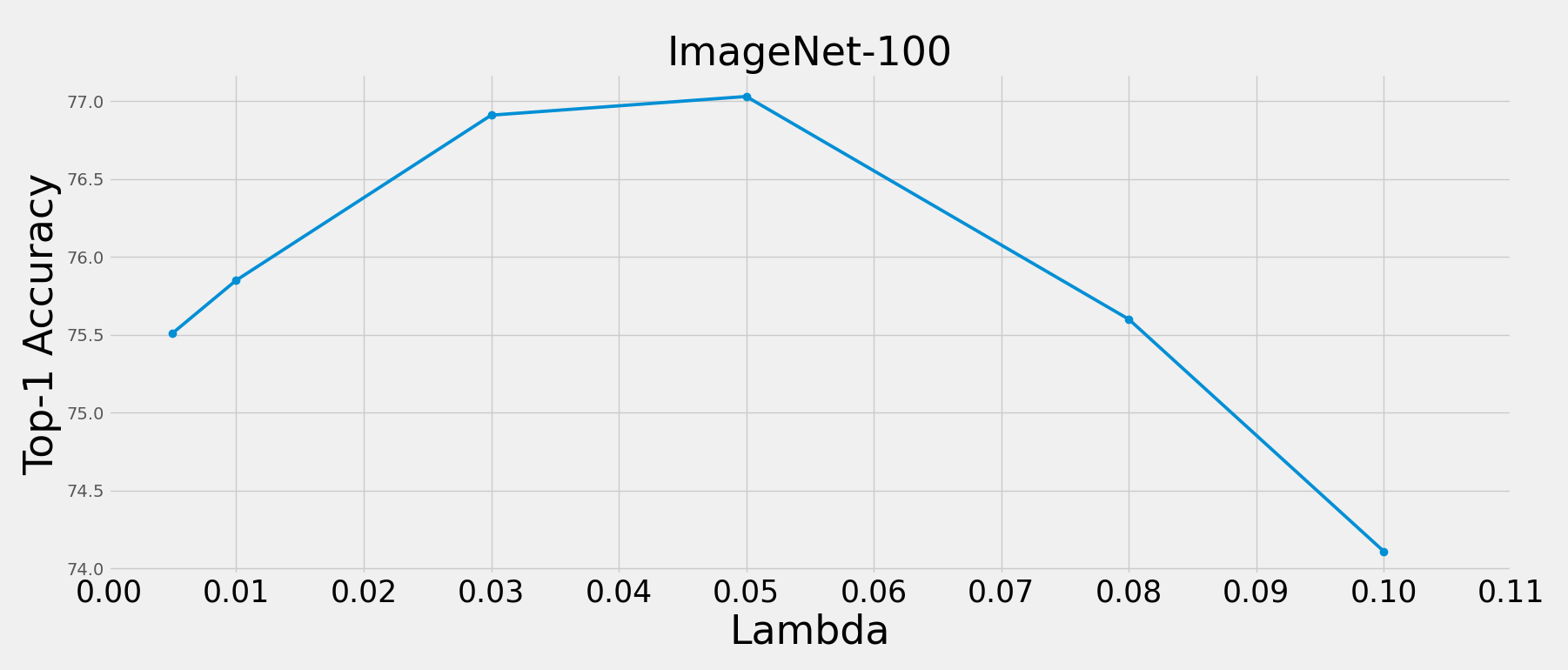}
		\caption{
		Sensitivity to $\lambda$.
\vspace{-1.5em}
}
\end{figure}

\subsection{Dynamic values for $\lambda$}
We also assess the case in which the value of $\lambda$ is not static, i.e., changing over time. The idea is to see the effect of stronger direct coloring in the early stages of the training, while it becomes less strong (relatively smaller $\lambda$) as the training progresses. Under this setting, we start with $\lambda=0.08$ and decrease it to $\lambda = 0.04$, scheduled as $[0.08,0.07,0.06,0.05,0.04]$, for epochs 1-50, 51-100, ..., 201-250. the top-1 accuracy is $76.70\%$, about $0.32\%$ less than the case with static $\lambda=0.05$.

\section{Simpler Design with Auto-correlation}
As discussed in the paper, we evaluated the direct coloring with a relatively simpler architectural design, substituting the cross-correlation with auto-correlation. Corresponding loss function for whitening, coloring and the total loss function are presented in this section. The total loss, coloring loss and whitening loss are as follows:
\begin{equation}
\label{eq1}
    \mathcal{L}= \mathcal{L}_{W'} + \lambda \mathcal{L}_{C'}
\end{equation}
where $\mathcal{L}_{C'}$ and $\mathcal{L}_{W'}$ are coloring loss and whitening loss, respectively. Here we have:
\begin{equation}
\label{eq2}
\begin{split}
     &\mathcal{L}_{C'} = \sum_i \sum_j(C'_{ij}-E'_{ij})^2\\
     &\mathcal{L}_{W'} = \sum_i(1-W'_{ii})^2+\alpha \sum_i \sum_{j\neq i} (W'_{ij})^2
     \end{split}
\end{equation}
where, $\lambda$ and $\alpha$ are weighting factors. Also $C'_{ij}$ and $W'_{ij}$ are elements of auto-correlation matrices computed for the coloring and whitening processes, respectively (see Fig. 3 in the paper), while $E'_{ij}$, is the target auto-correlation matrix used for the desired coloring. 
We also have:
\begin{equation}
    C'_{ij} \triangleq \frac{\sum_m' z_{m',i}z_{m',j}}{\sqrt{\sum_m' (z_{m',i})^2} \sqrt{\sum_m' (z_{m',j})^2}}
\end{equation}
 where $z$ is the normalized output of one coloring head for one view, $x_1$, and $m'$ is  the batch size (note that similar to the  original framework, here for each sample we fed two views to the network). Similarly, $W'{_{ij}}$ values are computed from the normalized output of one whitening head, where the auto-correlation is computed from all views fed to the network. Finally the elements of the matrix $E'$ is also computed from the latent space of one VAE similar to the equation for the elements of $C'$.

\section{More detail on Architecture}
The architectural design as well as some other details are presented here in more detail. Two architectures, ResNet18 and ResNet50 are used for encoder architecture, with the last layer replaced with a three layer projector, whitening projector. The last layer of the projector is the output with sizes 2048 and 1024. In fact in case of ResNet50 (under both linear and transfer learning settings) the output size is 2048 for ImageNet whereas in case of ResNet18 with other datasets, such as CIFAR10/100 the output size is 1024. Specifically the layer 16 and 46 of ResNet18 and ResNet50 pass through average pooling layer and then fed to the coloring heads (as conceptually shown in Fig. 1 of the paper). The architecture of the VAE is based on ResNet18 or ResNet50, and that determines the number of the layer that is connected to the coloring head. The coloring and whitening heads are made of three linear layers, first two are each followed by batch normalization and ReLU, whereas the output layer is fully connected layer of size 2048 or 1024. Finally, note that the layer that is connected to each coloring head first go through a max pooling process, similar to the layer that is connected to whitening head.

\section{Direct Coloring for other baselines}
In this section we evaluate the effectiveness of direct coloring for one other  baseline. We chose SIMSIAM as it is a modified version of BYOL, as a breakthrough work. We assess the case under standard augmentation as presented in the paper. The baseline accuracy without coloring head, under 1000 epochs of pre-training on ImageNet-100 using ResNet18 is $77.17\%$. Adding coloring heads, and corresponding term into the loss function, $\mathcal{L}=\mathcal{L}_{old}+\lambda\mathcal{L}_{Coloring}$, with $\lambda=0.01$, the top-1 accuracy upgrades to $78.40\%$, offering some $1.23\%$ improvement. Higher value of $\lambda$, $\lambda=0.05$, however, sharply degrades the accuracy to $72.5\%$. Hence, if used carefully, coloring can improve former baseline, SIMSIAM.


\section{Avoidance of complete collapse}
From theoretical perspective, we argue that any method that guarantees the avoidance of zero-variance representation, somehow assures the avoidance of complete collapse. In this sense, even whitening process also could  be considered as a helpful technique. Here we formulate the problem from the perspective of constraint optimization.
With direct coloring term added to the loss function of a gievn SSL framework which is prone to complete collapse, we have two terms (set of terms). First term being the original loss of the framework plus a second term corresponding to the direct coloring, $\mathcal{L}=f+\lambda g$. In essence, direct coloring in this setting would be seen as a constrained optimization problem. Thus, thinking in terms of Lagrange multipliers, one would see the first term alone, $f$,prone to a trivial solution, complete collapse. However, this is subjected to the second term, $g$, as the coloring constraint. We observe that  direct coloring would substantially reduce the chance of complete collapse, because to find the solution to this constrained optimization problem, the optimizer looks for points where the gradient vector of $f$ and $g$ are parallel to each other. Since complete collapse is only a solution to $f$ alone and certainly not a solution to $g$ (as $g$ encourages non-zero variance), if one chooses a proper $\lambda$, one can substantially avoid the complete collapse.

We assess it experimentally as well. The idea is to measure the representation variance, both in presence and without the presence of the coloring head. To this end, we measure the variance of last layers of the whitening head, output vector, as the coloring effect would be detectable there. Using the same architecture as Fig. 1 of the paper, this is done both with and without the coloring head (with $\lambda>0$ and $\lambda=0$). After 500 epochs of pre-training on ImageNet-100 with and without direct coloring, the weights are fixed and the variance of the normalized output vector of the whitening head is computed. In case of pre-training with direct coloring, the variance is $0.97$ while in case of pre-training without direct coloring, the variance is $0.68$, empirically confirming our theoretical analysis regarding the avoidance of complete collapse.

We performed the same experiments with the SIMSIAM, as the experimental setting is presented before. We measure the variance of the last layer of the projector head (a normalized vector), in presence and absence of the coloring head. The variance in presence of the coloring head is $0.89$ while in absence of the coloring heads the variance is $0.65$. This shows that coloring head in general would decrease the chance of complete collapse as it inject more variance to the representation, even with other baselines.

\end{document}